\documentclass[letterpaper, 10 pt, conference]{ieeeconf}  %

\IEEEoverridecommandlockouts                              %

\let\labelindent\relax

\title{\LARGE \bf
Dynamic Non-Prehensile Object Transport via \\ Model-Predictive Reinforcement Learning
}

\author{Neel Jawale$^{1}$, Byron Boots$^{1}$, Balakumar Sundaralingam$^{*2}$ and Mohak Bhardwaj$^{*1}$%
\thanks{$^{1}$ University of Washington $^{2}$ NVIDIA, USA, $^*$ equal advising}
}
\usepackage{hyperref}
\usepackage{graphics}
\usepackage{epsfig} 
\usepackage{xcolor}
\usepackage{amsmath} %
\usepackage{mathtools}
\usepackage{amssymb}  
\usepackage{enumitem}
\usepackage{comment}
\usepackage{tabularx}
\usepackage{algorithm}
\usepackage{algpseudocode}
\usepackage{multirow}
\usepackage{booktabs} %
\usepackage{graphicx} %
\usepackage{enumitem} %
\usepackage{hyperref} %
\usepackage{array}
\usepackage{float}
\usepackage{subcaption}
\usepackage{etoolbox}
\usepackage{cite}
\usepackage{xspace}
\usepackage{adjustbox}
\usepackage{placeins}
\usepackage{titlesec}
\titlespacing*{\subsection}
{0pt}{*0.7}{*0.6}

\definecolor{burntorange}{rgb}{0.8, 0.33, 0.0}
\definecolor{lightgreen}{rgb}{0.8,1,0.8}

\DeclareMathOperator*{\argmin}{arg\,min}

\newcommand{\argminprob}[1]{\underset{#1}{\argmin}}

\newcommand{\expect}[2]{\mathbb{E}_{#1}\left[#2\right]}

\newcommand{\bbm}{\begin{bmatrix}}
\newcommand{\ebm}{\end{bmatrix}}

\newcommand{\T}[2]{\prescript{#1}{}{T}^{#2}}
\newcommand{\twist}[2]{\prescript{#1}{}{v}^{#2}}
\newcommand{\spatialacc}[2]{\prescript{#1}{}{a}^{#2}}

\newcommand{\storm}{\textsc{STORM}\xspace}

\newcommand{\xxnote}[3]{}
\ifx\hidenotes\undefined
  \renewcommand{\xxnote}[3]{\color{#2}{#1: #3}}
\fi

\begin{document}

\setlength{\textfloatsep}{4pt}%

\setcounter{figure}{1}
\makeatletter
\let\@oldmaketitle\@maketitle%
\renewcommand{\@maketitle}{
   \@oldmaketitle%
   \begin{center}
    \centering      
    \includegraphics[width=0.95\textwidth]{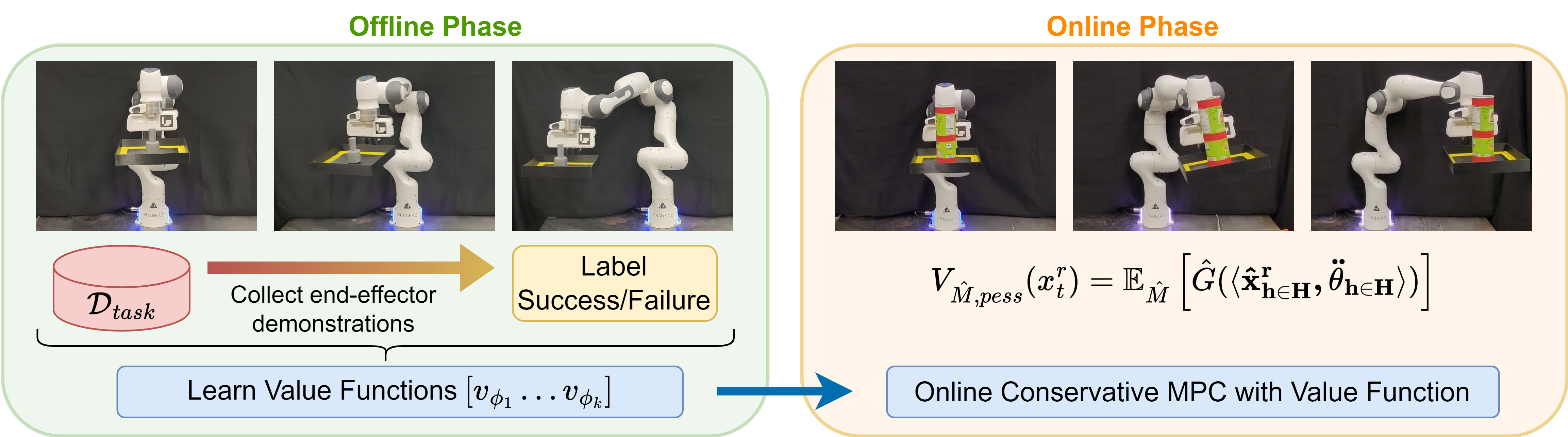}
   \label{fig:main_result}
  \end{center}
  \footnotesize{\textbf{Fig.~\thefigure:
   }~ The CV-MPC framework applied to the robot waiter problem. \textbf{(left)} In the offline phase, we collect end-effector only demonstrations of transporting object placed on the tray and label transitions with costs indicating slip ($c=c_{friction})$ or no-slip ($c=0$). We then train an ensemble of neural networks to independently predict the value function. \textbf{(right)} During online deployment, we use the pretrained value ensemble within a conservative MPC scheme to balance novel objects. 
  }\vspace{-12pt}
  \medskip}
\makeatother 

\maketitle
\thispagestyle{empty}
\pagestyle{empty}

\begin{abstract}
We investigate the problem of teaching a robot manipulator to perform dynamic non-prehensile object transport, also known as the `robot waiter' task, 
from a limited set of real-world demonstrations. We propose an approach that combines batch reinforcement learning (RL) with model-predictive control (MPC) by pretraining an ensemble of value functions from demonstration data, and utilizing them online within an uncertainty-aware MPC scheme to ensure robustness to limited data coverage. Our approach is straightforward to integrate with off-the-shelf MPC frameworks and enables learning solely from task space demonstrations with sparsely labeled transitions, while leveraging MPC to ensure smooth joint space motions and constraint satisfaction. 
We validate the proposed approach through extensive simulated and real-world experiments on a Franka Panda robot performing the robot waiter task and demonstrate robust deployment of value functions learned from 50-100 demonstrations. Furthermore, our approach enables generalization to novel objects not seen during training and can improve upon suboptimal demonstrations. We believe that such a framework can reduce the burden of providing extensive demonstrations and facilitate rapid training of robot manipulators to perform non-prehensile manipulation tasks.
Project videos and supplementary material can be found at: \url{https://sites.google.com/view/cvmpc}

\end{abstract}

\section{Introduction and Related Work}

\label{sec:introduction}

The ability to rapidly learn new tasks at deployment is a powerful skill for robot manipulators, especially when they need to operate in dynamic, contact-rich settings. Recent advances in imitation learning techniques based on behavior cloning (BC)
~\cite{mandlekar2020human, mandlekar2018roboturk} offer a promising avenue for training robot policies directly from a fixed set of task-specific demonstrations. 
However, a key shortcoming of BC approaches is covariate shift~\cite{ross2011reduction}, necessitating a large number of expert demonstrations to generalize to different scenarios that the robot encounters. This problem is aggravated in tasks requiring dynamic interactions where it is hard to recover from mistakes and the robot must be robust to dynamics and perception uncertainty. Recent works such as~\cite{zhaoaloha} have demonstrated that imitation learning with large scale data collection can scale to complex, dynamic manipulation tasks. However, for general practitioners collecting extensive demonstrations can be prohibitively time consuming and expensive, thus requiring approaches that can learn efficiently from limited datasets. \looseness=-1       

In this work, we present an approach for teaching robot manipulators to perform dynamic manipulation tasks from a small number of real-world demonstrations. In particular, we focus on the task of non-prehensile object transport, also known as the classic `robot waiter' problem~\cite{selvaggio2022nonpreobjtransport}. In this task, the robot must dynamically transport an object placed on a tray grasped by its end-effector to different workspace locations while ensuring object stability and preventing slip as shown in Fig.~1. This task is representative of a large class of non-prehensile manipulation scenarios where the robot cannot directly constrain object motion through stable grasping and must demonstrate precise motions under uncertainty. This setting makes providing extensive demonstrations extremely challenging. Furthermore, the provided demonstrations can be suboptimal and robot must be able to generalize to novel objects not seen during training.  

Recent advances in joint-space model-predictive control (MPC) for robotic manipulators have made optimizing over costs in task-space (reaching a pose, avoiding collisions) and joint-space (joint limits, singularities) practical, particularly with the introduction of GPU accelerated frameworks such as~\cite{bhardwaj2022storm, pezzato2023samplingbased}. Motivated by the availability of such tools, we explore how we can leverage MPC implementations to quickly train robots to perform the dynamic non-prehensile object transport task from a small set of real-world demonstrations while ensuring safe deployment on hardware. Furthermore, we also consider a setting where the robot manipulator is provided with only task space (i.e end-effector) demonstrations containing sparse cost information (slip/no-slip), that can significantly reduce demonstrator burden for dynamic manipulation tasks. \looseness=-1 

We propose Conservative Value MPC~(CV-MPC), a framework based on infinite-horizon MPC~\cite{lowrey2018plan, erez2012infinite, bhardwaj2020blending} that combines value functions learned from data with model-based control for sample efficient learning of dynamic manipulation tasks.
Our approach consists of two phases, as shown in Fig.
1: (1) \textit{Offline Value Function Learning} where we train an ensemble of value functions on demonstrated trajectories to independently approximate the long-term probability of failure of the demonstrator, from only end-effector observations, and (2) \textit{Online Conservative MPC} where the learned ensemble is used to construct an uncertainty-aware approximation of the trajectory returns to explicitly penalize the controller from visiting out of distribution states. This scheme is motivated by offline RL methods~\cite{levine2020offline, kumar2020conservative} aimed to tackle covariate shift due to missing data coverage. We discuss the comparison to related approaches in literature in Sec.~\ref{sec:approach:pess_mpc}. Our return estimation scheme is straightforward to integrate with existing off-the-shelf MPC frameworks such as~\cite{bhardwaj2022storm,pezzato2023samplingbased}, allowing users to leverage these powerful algorithms to achieve feasible joint-space motions while generalizing to novel situations. We hypothesize that such a setup can scale beyond the robot waiter task and enable end users to teach robots new dynamic skills via teleoperation while having minimal knowledge about robotics or reinforcement learning. 
In summary, our main contributions are 
\begin{enumerate}
    \item A framework for learning the dynamical non-prehensile robot waiter task from only task space demonstrations with sparse labels indicating success or failure.  
    \item A novel approach combining offline RL with online MPC to learn dynamic manipulation tasks from limited demonstrations while ensuring robust deployment on real hardware.
    \item Exhaustive empirical evaluations with over 800 trials across 13 objects on a real-world Franka Panda arm demonstrating generalization to novel objects, improving over sub-optimal experts, all from 50-100 demonstrations.  
\end{enumerate}

\section{Problem Setup}
\label{sec:task}

We define the robot waiter problem as generating actions for a robot manipulator to transport a grasped tray containing an object to target workspace locations while preventing the object from slipping. This task requires the robot to reason about the contact dynamics between the object and the tray in the cartesian-space while optimizing for control inputs in the joint-space in real time. As discussed in Sec.~\ref{sec:introduction}, we explore solving this problem via model predictive control~(MPC) motivated by recent successes~\cite{abbeel2010autonomous, williams2016aggressive, di2018dynamic}.
We choose actions to be desired joint accelerations~$\ddot{\theta}_t$, integrated to obtain desired joint velocity and position~$\dot{\theta}_t, \theta_t$ targets respectively. We use forward kinematics to obtain the state of the end-effector~(gripper) as $x^r_t = [\T{w}{ee}, \twist{w}{ee}, \spatialacc{w}{ee} ]$ where $\T{w}{ee}$ is the SE(3) pose transform of the end-effector in world frame $w$, $\twist{w}{ee}$ and $\spatialacc{w}{ee}$ are the spatial twists and accelerations respectively. We assume that the tray is rigidly grasped in a fixed, known pose  w.r.t. the gripper. 

There is a rich history of prior works that have successfully applied MPC to the robot waiter problem 
by formulating cost functions that encourage task completion
~\cite{selvaggio2022nonpreobjtransport, heins2023keep, 9812424}. However, these approaches rely on analytical formulations of contact dynamics with accurate knowledge of object geometry, dynamics, and friction properties. This can be a strong assumption in many real-world scenarios where biased estimates of inertial and frictional properties of the object can lead to poor task success as we demonstrate in our results in Sec.~\ref{sec:sim_exp}.
We propose to overcome these issues by augmenting MPC with value functions learned from successful and failed task demonstrations. 

\textit{Learning from Task-Space Demonstrations:} We consider a setting where the robot is provided with a fixed dataset of task-space demonstrations ($\mathcal{D}_{\text{task}}$) containing transition tuples $(x^r_t, c_t, x^r_{t+1})$ where $x^r_t$ is the end-effector state.
Importantly, $\mathcal{D}_{\text{task}}$ does not contain the robot joint state ($\theta_t, \dot{\theta_t}$) or demonstrated joint-space actions $\ddot{\theta_t}$, 
formally making this an instance of offline policy learning from observations alone~\cite{li2023mahalo}.
Further, since annotating every transition with a dense cost label can be an arduous process, we consider a more flexible setting of learning with sparse cost labels denoting states that lead to task failure. In the robot waiter task, this corresponds to simply annotating end-effector motions that lead to the object slipping. Finally, while demonstrations are provided in task space, the robot is expected to produce smooth joint-space motions while respecting constraints like joint limits, singularity avoidance, and collision avoidance. \looseness=-1   
Next, we detail our two-stage approach to address these challenges: offline value function learning from observations (Sec.~\ref{sec:offline_learning}) followed by using the learned value function online with a conservative MPC (Sec.~\ref{sec:online_mpc}). 

\section{Approach}\label{sec:approach}
\subsection{Offline Value Function Learning}
\label{sec:offline_learning}
We model the robot waiter task as learning a control policy in a Markov Decision Process (MDP)~\cite{sutton2018reinforcement}. In this setting, a value function approximates the expected cost-to-go (or return) of a policy. We denote a trajectory of robot states and controls by $\langle\mathbf{x^r_t}, \mathbf{\ddot{\theta}_t}\rangle$ and  $G(\langle\mathbf{x^r_t}, \mathbf{\ddot{\theta}_t}\rangle) = \sum_{t'=t}^{\infty} \gamma^{t-t'} c(x^r_{t'}, \ddot{\theta}_{t'})$ as its total return with discount factor $\gamma \in [0, 1)$. The value function is given by $V(x^r_t) = \expect{}{G(\langle\mathbf{x^r_t}, \mathbf{\ddot{\theta}_t}\rangle)}$ where the expectation is over dynamics and control policy. 

During the offline phase, we train an ensemble of $K$ value functions $\left[ V_{\phi_1}(x^r_t), \ldots ,V_{\phi_K}(x^r_t) \right]$ parameterized by $\left[\phi_1 \ldots \phi_K \right]$ 
to approximate the cost-to-go for the end-effector trajectories in $\mathcal{D}_{\text{task}}$. 
Each ensemble member $i \in [1 \ldots K]$ is independently initialized and trained on mini-batches of data to optimize the Bellman error objective~\cite{bertsekas1996neuro}
\begin{equation}
\label{eq:value_fn_loss}
    \phi^{*}_{i} = \argminprob{\phi}{\;\expect{(x, c, x') \sim \mathcal{D}_{task}}{\;(c + \gamma V_{\phi} (x') - V_{\phi}(x))^2 }}
\end{equation}
The rationale behind learning an ensemble of value functions is to capture the epistemic uncertainty arising from limited data coverage, similar to prior ensemble-based approaches in model-based~\cite{kidambi2020morel} and model-free~\cite{an2021uncertainty} offline RL.
In the next step, we leverage this uncertainty to construct a conservative value function estimate for online MPC. 
\subsection{Conservative Value MPC (CV-MPC)}\label{sec:approach:pess_mpc}
\label{sec:online_mpc}
MPC has a rich history in feedback control of complex robotic systems under dynamic constraints 
\cite{abbeel2010autonomous, williams2016aggressive, di2018dynamic}. 
Instead of searching for a global policy, 
MPC optimizes local policies online at a given state $x^r_t$ using an approximate model ($\hat{M}$) with a finite lookahead horizon $H$, by minimizing a model-based approximation of the value function 
$V_{\hat{M}}(x^r_t) = \expect{\hat{M}}{\hat{G}(\langle\mathbf{\hat{x}^r_{h \in H}, \mathbf{\ddot{\theta}_{h \in H}}} \rangle)}$ where 
$\hat{G} (\langle\mathbf{\hat{x}^r_{h \in H}, \mathbf{\ddot{\theta}_{h \in H}}} \rangle) = \sum_{t'= 0}^{H-1} \gamma ^{t'} \hat{c} (\hat{x}^r_{t'}, \ddot{\theta_{t'}}) + \gamma^H \hat{V}(\hat{x}^r_{H})$ is a model-based estimate of the trajectory return, with $\hat{V}$ being a terminal cost-to-go estimate~\cite{bhardwaj2020blending}.

During online execution, we use the pretrained value function ensemble to approximate trajectory returns as part of the MPC objective to optimize for control inputs that can achieve non-prehensile transport task. We assume that MPC has a sufficiently accurate robot model to predict end-effector states resulting from applied controls, a reasonable assumption in most robotics tasks. However, due to limited coverage of $\mathcal{D}_{task}$ the learned value functions can be susceptible to extrapolation errors when MPC queries them in different parts of the state space. Thus, naively using them to predict trajectory returns can lead to divergence in local policies optimized by MPC. To address this, we take inspiration from offline RL methods~\cite{kumar2020conservative, cheng2022adversarially} that explicitly constrain the policy from visiting parts of state space lying outside data support by optimizing pessimistic (or conservative) upper bounds of the value function (lower bound in reward formulation). Let $\hat{G}_{i}(\mathbf{\hat{x}^r_{h \in H},  \mathbf{\ddot{\theta}}_{h \in H}}) = \sum_{t'=t}^{t+H} \gamma ^{t'-t} \hat{V_{\phi,i}} (\hat{x}^r_{t'})$ denote an estimate of the $H$ step return of a simulated trajectory as predicted by ensemble member $V_{\phi,i}$. 
We introduce pessimism in MPC updates by employing a risk-averse objective \looseness=-1
\begin{equation}
\label{eq:pess_returns}
    \hat{G}_{pess}(\mathbf{\hat{x}^r_{h \in H}, \mathbf{\ddot{\theta}_{h \in H}}}) = \log (\sum_{i=1}^{K} \exp(\frac{1}{\lambda} \hat{G}_{i}(\mathbf{\hat{x}^r_{h \in H},  \mathbf{\ddot{u}_{h \in H}}}) )
\end{equation}

MPC then optimizes a corresponding conservative value estimate $V_{\hat{M}, pess}(x^r_t) = \expect{\hat{M}}{\hat{G}(\langle\mathbf{\hat{x}^r_{h \in H}, \mathbf{\ddot{\theta}_{h \in H}}}\rangle )} $. Intuitively, the above estimator forces MPC to downweight trajectories that lead to regions of state space where the ensemble members have higher disagreement with $\lambda$ being the hyper-parameter controlling the amount of pessimism.
It is important to note that we compute the pessimistic estimate only at the initial state $x_t$ (initial state pessimism), and not at all intermediate states in MPC rollouts (point-wise pessimism), which can lead to an over-pessimism bias~\cite{xie2021bellman}.  

Our pessimisitc value estimate involves a simple modification to the MPC objective, and is straightforward to integrate with existing algorithms as another cost function term. 
This enables us to leverage powerful off-the-shelf MPC frameworks to ensure constraint satisfaction, motion quality and safety during deployment. In this work, we use the GPU accelerated sampling-based MPC framework called \storm from~\cite{bhardwaj2022storm}, that has demonstrated great performance in reactive control tasks, by augmenting their trajectory returns (denoted $\hat{G}_{STORM}$) with $\hat{G}_{pess}$ to get $\hat{G} = \hat{G}_{STORM} + \hat{G}_{pess}$. \footnote{Refer to~\cite{bhardwaj2022storm} for details about the STORM framework.}

\noindent\textbf{Comparison with other approaches:} Prior approaches in literature
\cite{sikchi2022learning, argenson2020model} have also proposed using offline RL to pretrain terminal value functions for MPC, however, the pretraining is usually done via pointwise pessimistic algorithms i.e training value functions to be pessimistic for all states, which is shown to have an over-pessimism bias~\cite{xie2021bellman}. In CV-MPC, 
the value functions themselves are not pessimistic, but conservatism is introduced at trajectory level only during online MPC updates. This allows each learned function to independently approximate the cost-to-go which we found to help induce diverse predictions. We also use a sum of value predictions for the returns in Eq.~\ref{eq:pess_returns} instead of just a terminal value prediction. This is motivated by~\cite{bhardwaj2020blending} where it was shown that blending value estimates from real-world data at all steps in the horizon can help mitigate model errors. \looseness=-1    

\subsection{Demonstrations} 
In this work, we collect demonstration data using an MPC based algorithmic demonstrator based on the formulation from~\cite{heins2023keep}. This formulation uses known geometric, inertial, and friction properties of the object to compute friction cone constraints that are used within an MPC optimizer to solve for motions that ensure object does not slip. We refer the reader to Sec \ref{sec:mpcdemonstrator} for a detailed description of the MPC demonstrator. 
It is important to note that we only use end-effector trajectories for learning, meaning that while the demonstrator has access to the true physical properties of the object, the learner does not. While our approach does not make any assumptions about how demonstrations are collected,   
utilizing an algorithmic demonstrator, 
reduces data collection time in simulation and real-world, and provides a reliable baseline for evaluation. 
However, our approach is equally compatible with other forms of demonstrators such as human teleoperation.

\section{Experimental Evaluation}\label{sec:experiments}
We validate our approach using extensive simulated and real-world experiments with a Franka Panda robot performing the waiter task. Specifically, we design experiments to test whether our approach can (i) learn in a sample efficient manner, (ii) demonstrate agile behaviors while satisfying constraints, (iii) generalize to diverse objects not seen during training, and (iv) improve over a suboptimal demonstrator. All experiments are performed on a single RTX 3090 GPU, and learned value functions are deployed within the proposed CV-MPC approach on the robot at a control frequency of 50 Hz.

\noindent\textbf{Setup:}
 For all experiments, we set the number of training demonstrations ($\mathcal{D}_{\text{task}}$) to 50, except for case study 2, which requires 100 demonstrations to achieve better generalization. Each set of 50 demonstrations takes approximately 5 minutes to collect in simulation and 10 minutes in the real-world due to additional reset time. Goal positions are randomly sampled within the workspace of the robot during both demonstrations and evaluation. The input to the value function is only the end-effector state described in Sec.~\ref{sec:task}.
 The value function does not learn to reach target positions for the tray, instead we use an additional L2 norm cost from~\cite{bhardwaj2022storm}. 

\noindent\textbf{Demonstration Data:}
We collect a dataset $\mathcal{D}_{task}$ of demonstrations containing only end-effector states for balancing an object across different goal locations (Sec.~\ref{sec:task}).
We label the transitions in the dataset with a cost indicating whether the end-effector motions result in violating object slip constraints. 
At timestep $t$, the cost label $c(x^r_t) \in \lbrace 0, {c}_{friction} \rbrace$ where ${c}_{friction}$ represents the friction cone violation resulting in object slip.
In simulation, we automatically label demonstrations as success or failure based on the slip distance of the object. 
For real-world experiments, demonstrations are labeled by a human operator monitoring the object slip in real-time using the markings on the tray. We terminate a demonstration episode when the object slips by 2 cm on the tray.

\noindent\textbf{Performance Metrics:} 
We consider a trial successful if the end-effector reaches the goal position within a 2 cm margin, with the L2 norm of the combined linear and angular velocities constrained to less than 0.02, ensuring both low linear velocity (below 0.02 m/s) and low angular velocity (below 0.02 rad/s), while avoiding object slip. %
For object slip evaluation, \textbf{success} is defined as the object moving less than 2 cm on the tray after reaching its goal position from the initial state. 
Since moving at higher velocities allows the robot to tilt the tray more while ensuring object stability, we measure the dynamic characteristics of the motion by calculating the \textbf{maximum tilt angle (\(\alpha_{\mathrm{ee}, t}^{\max}\))}, the \textbf{maximum linear velocity norm (\(\|\mathbf{v}_{\mathrm{ee}, t}\|_{\max}\))}, and the \textbf{maximum angular velocity norm (\(\|\boldsymbol{\omega}_{\mathrm{ee}, t}\|_{\max}\))} achieved across all successful episodes. 

\subsection{Simulation Experiments}
\label{sec:sim_exp}
We conduct several ablation studies in simulation to identify key hyper-parameters and validate the efficacy of our approach.~\looseness=-1

\noindent\textbf{Pessimism Parameter ($\lambda$) - Ensemble Size ($K$) Ablation:}
We first study the impact of two key parameters on the performance of learned value functions in CV-MPC: the pessimism parameter ($\lambda$) and ensemble size ($K$). We also compare the effectiveness of learning a value function versus a simple one-step cost function in MPC.

\begin{figure}
    \centering \includegraphics[width=0.90\linewidth]{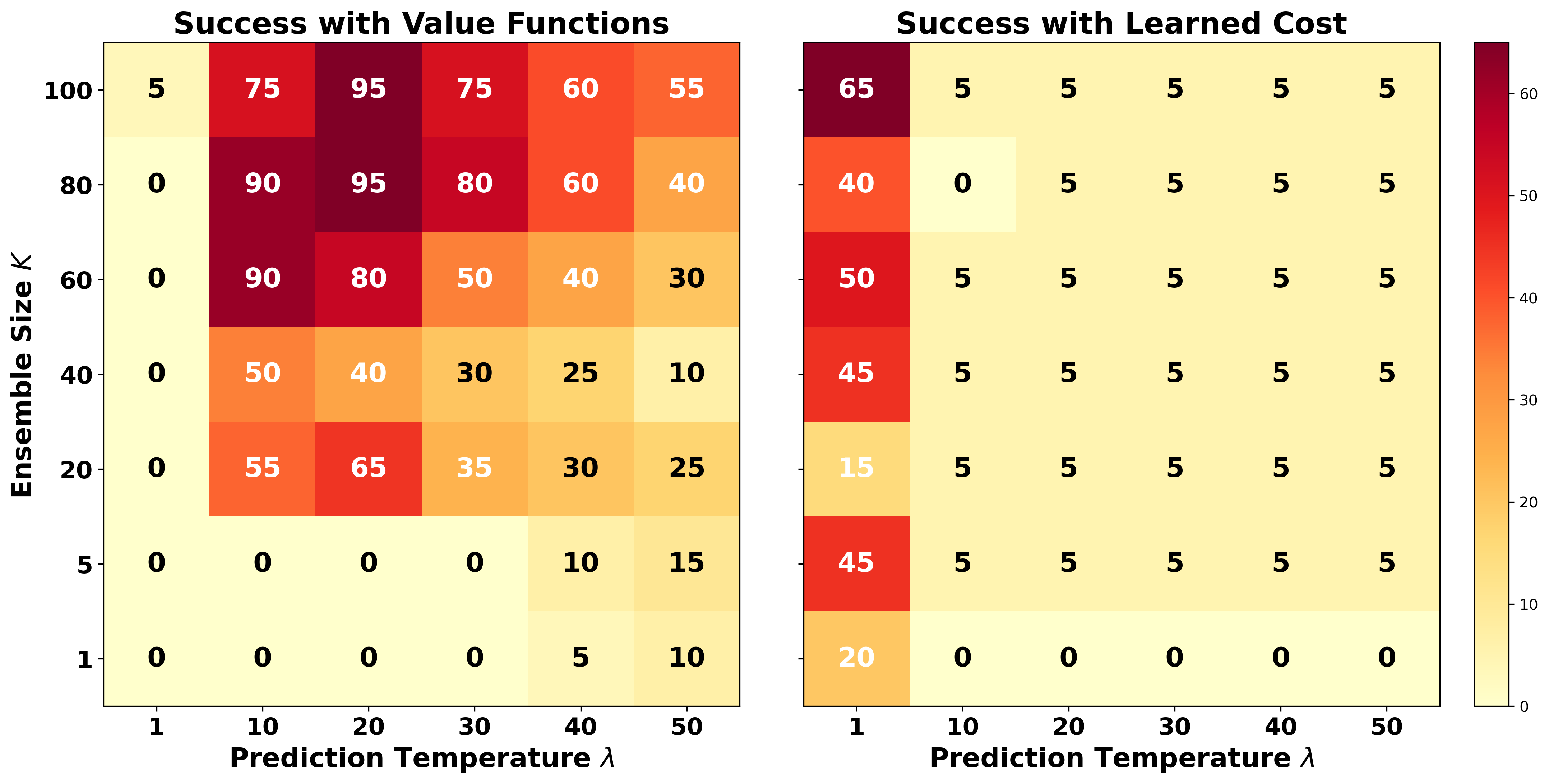}
    \caption{(Simulation experiment) Comparison of success rates across ensemble size $K$ and pessimism $\lambda$ for (a) Value Functions and (b) Learned One-Step Cost. For Value Functions, $\lambda = 20$ and $K = 80$ yield the best performance, though results are robust across a range of parameter combinations, reducing the need for extensive fine-tuning. In contrast, the Learned One-Step Cost exhibits lower overall success, with relatively high performance limited to $\lambda = 1$, highlighting its difficulty in handling sparse rewards.}
    \label{fig:success_comparisons_costs}
\end{figure}

\begin{figure*}[ht]
    \centering
    \begin{subfigure}[b]{0.32\textwidth}
        \centering
        \includegraphics[width=\textwidth]{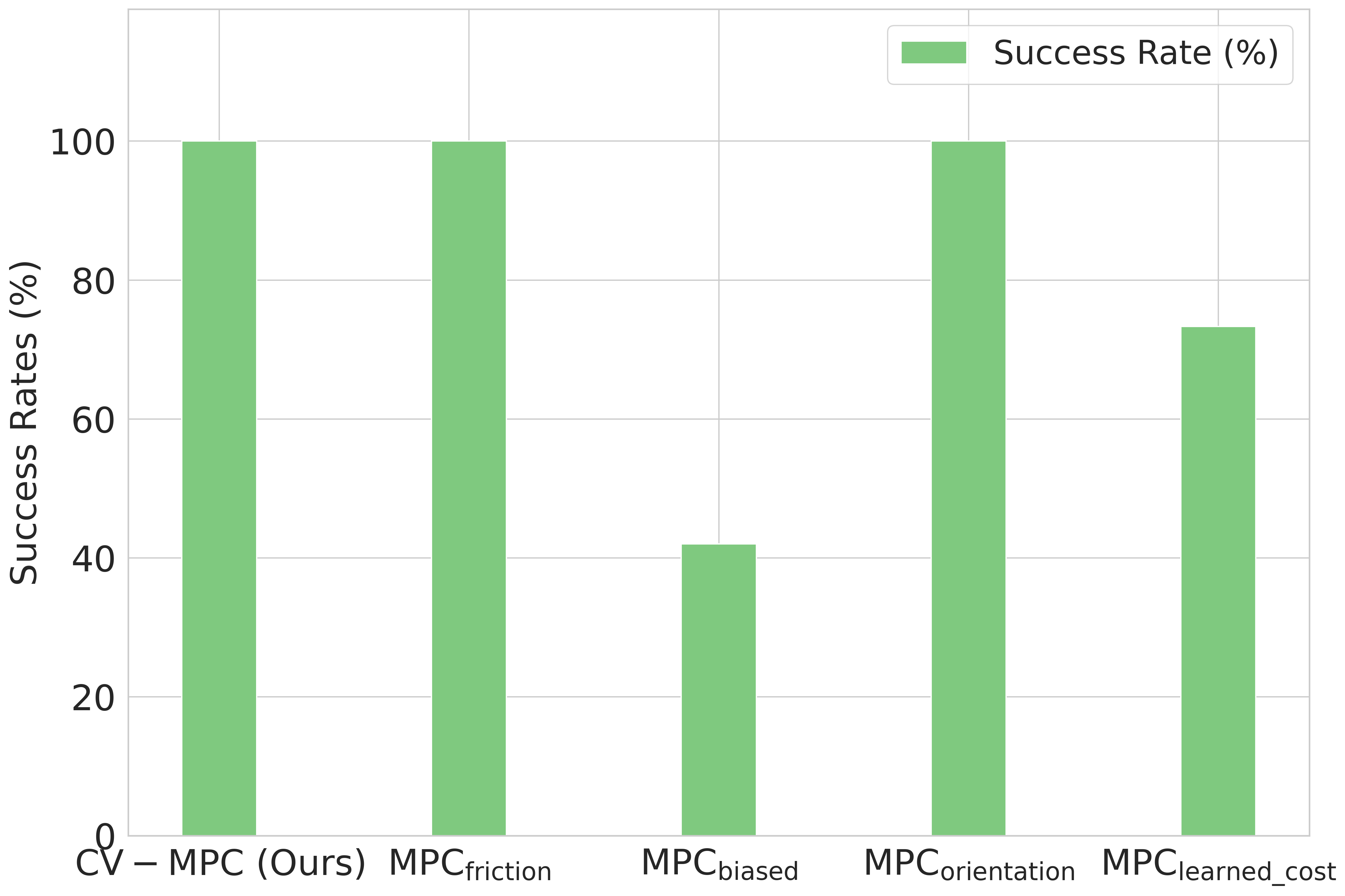}
        \label{fig:success_rates}
    \end{subfigure}
    \hfill
    \begin{subfigure}[b]{0.32\textwidth}
        \centering
        \includegraphics[width=\textwidth]{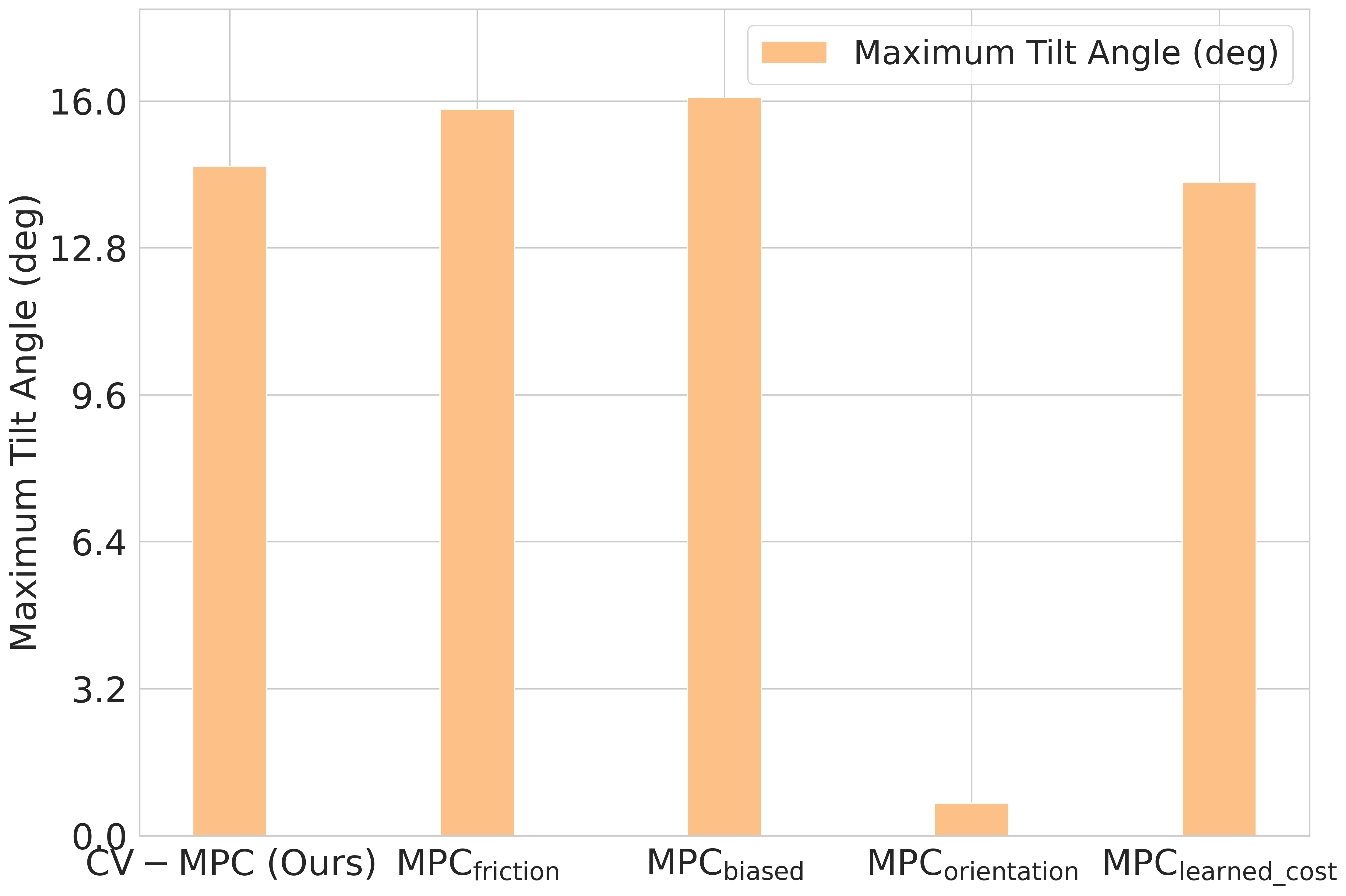}
        \label{fig:alpha}
    \end{subfigure}
    \hfill
    \begin{subfigure}[b]{0.32\textwidth}
        \centering
        \includegraphics[width=\textwidth]{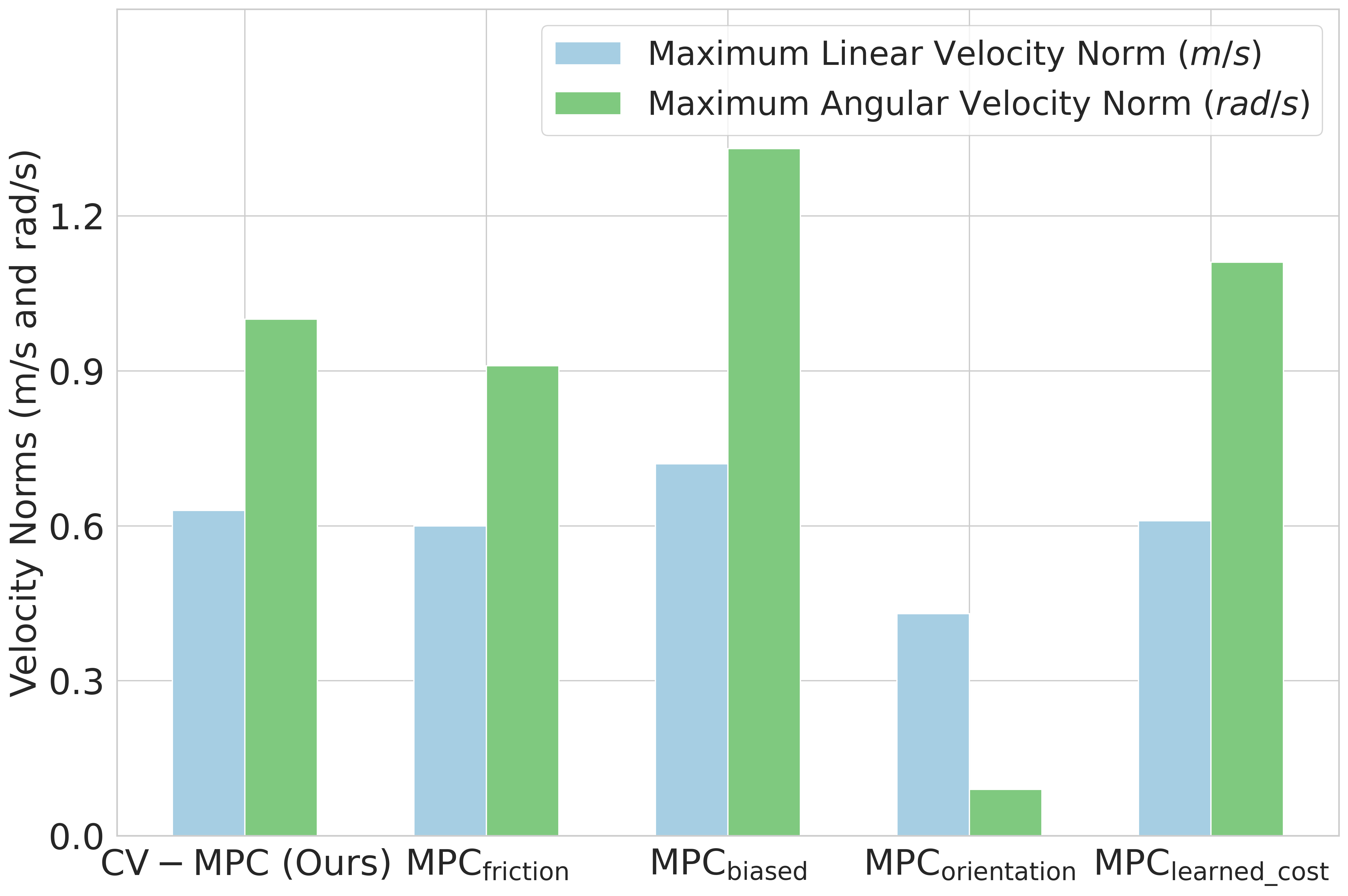}
        \label{fig:vels}
    \end{subfigure}
    \vspace{-0.1cm}
    \caption{(Simulation Experiment) Comparison of bar plots for performance metrics evaluated across 60 trials per algorithm between our proposed CV-MPC approach and several baselines: {MPC}$_{\text{friction}}$, a demonstrator with accurate knowledge of object properties; {MPC}$_{\text{biased}}$, a biased demonstrator assuming incorrect object properties (friction coefficient $\mu$); {MPC}$_{\text{orientation}}$, which employs a high orientation-maintaining cost; and {MPC}$_{\text{learned\_cost}}$, which learns the one-step friction cost from demonstrations instead of utilizing a value function. The CV-MPC approach performs comparably to the demonstrator with true object properties and surpasses the other baselines, either in terms of success or dynamic behavior.}
    \label{fig:baselines}
    \vspace{-0.2cm}
\end{figure*}

\noindent\textit{a) Value Function Ablations:}
In order to analyze the effects of ensemble size ($K$) and pessimism ($\lambda$) on value function performance, we use a cube with $\mu=0.3$, mass of 0.05 kg, and side length of 0.05 m. We sample random workspace goals for the robot to reach while maintaining object stability and test the success rate for different combinations of $K$ and $\lambda$ (20 episodes per hyper-parameter combination). The resulting heatmaps are visualized in Figure~\ref{fig:success_comparisons_costs}.
The results indicate that CV-MPC is remarkably robust to hyper-parameter settings, suggesting the feasibility of real-world deployment without extensive fine-tuning. For the remainder of the experiments, we use the best-performing setting of ($K=80, \lambda=20$). 

\noindent\textit{b) Learning One-step Cost vs. Value Function:}
We study if the long-horizon reasoning of value functions is essential for performing the object transport task. To do so, we consider  
a baseline of MPC with a learned one-step cost, and conduct the same ablation as above  
to see if increasing ensemble size improves learning from sparse labels. 
As shown in Figure~\ref{fig:success_comparisons_costs}, an ensemble size of 100 achieves the highest success rate; however, the success drops sharply for pessimism values other than 1, showing the sensitivity of this method to $\lambda$. This sensitivity stems from the difficulty of learning cost functions from sparse labels. Higher $\lambda$ increases pessimism, causing more frequent failures. In contrast, value functions excel due to their long-term state evaluation: excessive pessimism enhances stability, while too little leads to overly optimistic and fragile strategies.

\noindent\textbf{Comparison to Baselines:} We compare CV-MPC against several baselines: Demonstrator {MPC}$_{\text{friction}}$ ~(Sec. \ref{sec:approach}), that applies MPC with a friction cost; {MPC}$_{\text{biased}}$, that also uses MPC with friction cost but assumes incorrect knowledge of friction coefficient $\mu$; {MPC}$_{\text{orientation}}$, that incorporates a high orientation cost; and {MPC}$_{\text{learned\_cost}}$, that relies on a learned one-step cost function derived from demonstrations, as illustrated in Fig.~\ref{fig:baselines}. We apply these methods to the same object (cube) used in the previous ablation experiment. Our approach performs comparably to the demonstrator, which has precise knowledge of object properties and displays similar dynamic behavior. %
Importantly, CV-MPC significantly outperforms the demonstrator when the demonstrator operates with an incorrect friction value. While {MPC}$_{\text{orientation}}$ achieves a high success rate, it is notably slower and conservative.
In contrast, {MPC}$_{\text{learned\_cost}}$ performs worse than our method due to the sparsity of the cost function and no long-horizon reasoning.

\begin{figure}
    \centering    \includegraphics[width=0.90\linewidth]{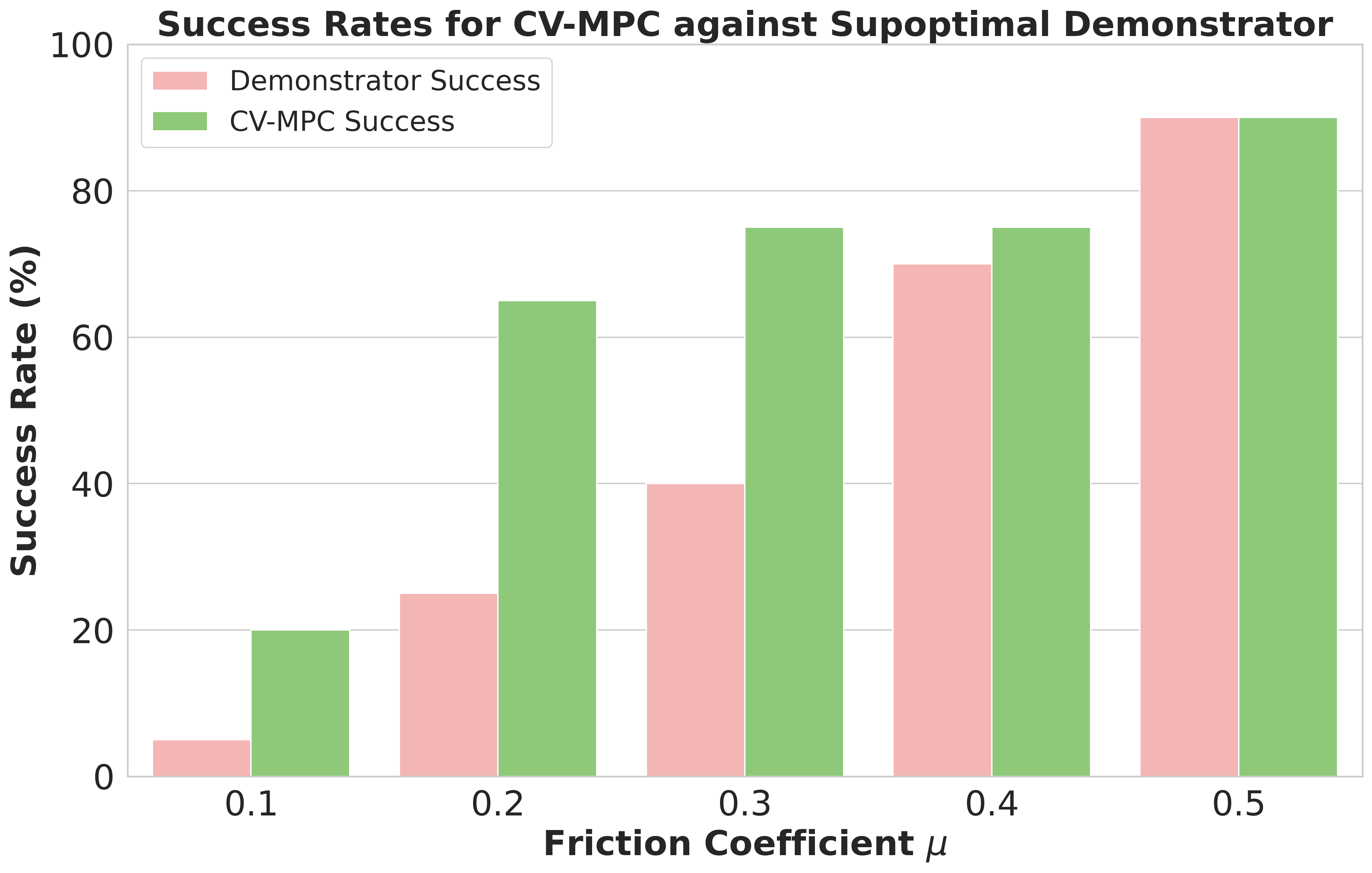}
    \caption{(Simulation Experiment) Success comparison of the demonstrator with a biased value of friction coefficient $\mu$ vs. CV-MPC evaluated across 20 trials per $\mu$. Expert has higher $\mu = 0.6$ than true values shown in different columns. By learning value functions across different settings, CV-MPC can improve over the suboptimal demonstrator.}
    \label{fig:biased_demo_plot}
\end{figure}

\noindent\textbf{Improving Over Sub-optimal Demonstrators:}
In dynamic real-world settings, it is often hard to provide expert demonstrations and it is essential to learn from examples of failure. We study if CV-MPC can  improve over suboptimal demonstrations via offline RL. We consider learning from demonstrations given by the MPC demonstrator with incorrect knowledge of the friction coefficient ($\mu = 0.6$), which is generally hard to estimate in real-world settings. We evaluate the demonstrator in scenarios where the true $\mu < 0.6$ and learn value functions for each scenario from the suboptimal demonstrations. Please note that we use the same cube here as the previous experiments but vary the $\mu$. From Fig.~\ref{fig:biased_demo_plot}, we see that as we decrease the friction coefficient, the success of CV-MPC is higher than the demonstrator. This shows that our ensemble of value functions effectively learns from the demonstrator's failures and biases, thereby improving on it. 

\begin{table*}[bt]
\centering
\begin{adjustbox}{max width=1\textwidth}
\renewcommand{\arraystretch}{1.5} %
\normalsize
\begin{tabular}{|c|>{\centering\arraybackslash}m{2.5cm}|>{\centering\arraybackslash}m{2.5cm}|>{\centering\arraybackslash}m{2.5cm}|>{\centering\arraybackslash}m{2.5cm}|>{\centering\arraybackslash}m{2.5cm}|>{\centering\arraybackslash}m{2.5cm}|>{\centering\arraybackslash}m{2.5cm}|>{\centering\arraybackslash}m{2.5cm}|>{\centering\arraybackslash}m{2.5cm}|}
\hline
\textbf{\large Metrics:} & 
\textbf{\large Training Object} & 
\textbf{\large Test Object \#1} & 
\textbf{\large Test Object \#2} & 
\textbf{\large Test Object \#3} & 
\textbf{\large Test Object \#4} & 
\textbf{\large Test Object \#5} & 
\textbf{\large Test Object \#6} & 
\textbf{\large Test Object \#7} & 
\textbf{\large Test Object \#8} \\
\hline
\large \textbf{$Success$ (\%)} & \large 95.00 & \large 88.33 & \large 73.33 & \large 86.67 & \large 88.33 & \large 83.33 & \large 85.00 & \large 78.33 & \large 81.67  \\ 
\hline
\large \textbf{${\alpha_{\mathrm{ee}, t}^{\max}}$ (deg)} & \large 16.41 ± 0.12 & \large 13.29 ± 0.20 & \large 12.65 ± 0.37 & \large 15.47 ± 0.70 & \large 13.43 ± 0.97 & \large 14.94 ± 0.82 & \large 15.30 ± 0.32 & \large 12.31 ± 0.39 & \large 12.47 ± 0.43  \\ 
\hline
\large \textbf{$\|\mathbf{v}_{\mathrm{ee}, t}\|_{\max}$ (m/s)} & \large 0.60 ± 0.01 & \large 0.57 ± 0.03 & \large 0.53 ± 0.04 & \large 0.56 ± 0.02 & \large 0.57 ± 0.01 & \large 0.56 ± 0.02 & \large 0.51 ± 0.02 & \large 0.51 ± 0.03 & \large 0.57 ± 0.02  \\ 
\hline
\large \textbf{$\|\boldsymbol{\omega}_{\mathrm{ee}, t}\|_{\max}$ (rad/s)} & \large 0.91 ± 0.03 & \large 0.88 ± 0.03 & \large 0.77 ± 0.04 & \large 0.86 ± 0.03 & \large 0.85 ± 0.03 & \large 0.90 ± 0.03 & \large 0.87 ± 0.03 & \large 0.74 ± 0.03 & \large 0.77 ± 0.03  \\ 
\hline
\end{tabular}
\end{adjustbox}
\vspace{0.1cm}
\caption{(Real-World Experiment) Performance (mean ± std. error for metrics except for success) and generalizability of CV-MPC on convex-shaped solid objects reported across 60 trials per object. CV-MPC demonstrates superior performance when applied to the cube used during the collection of demonstrations. Notably, the method also achieves comparable results on previously unseen convex-shaped objects, exhibiting only a marginal decline in the success metric.}
\label{tab:gen1}
\end{table*}

\subsection{Real-World Experiments}
\label{sec:real_exp}
From our real-world experiments on a Franka Panda robot we test how CV-MPC can enable robust deployment of learned value functions in the real-world and generalization abilities of our proposed setup.
\begin{figure}%
    \centering    \includegraphics[width=1\linewidth]{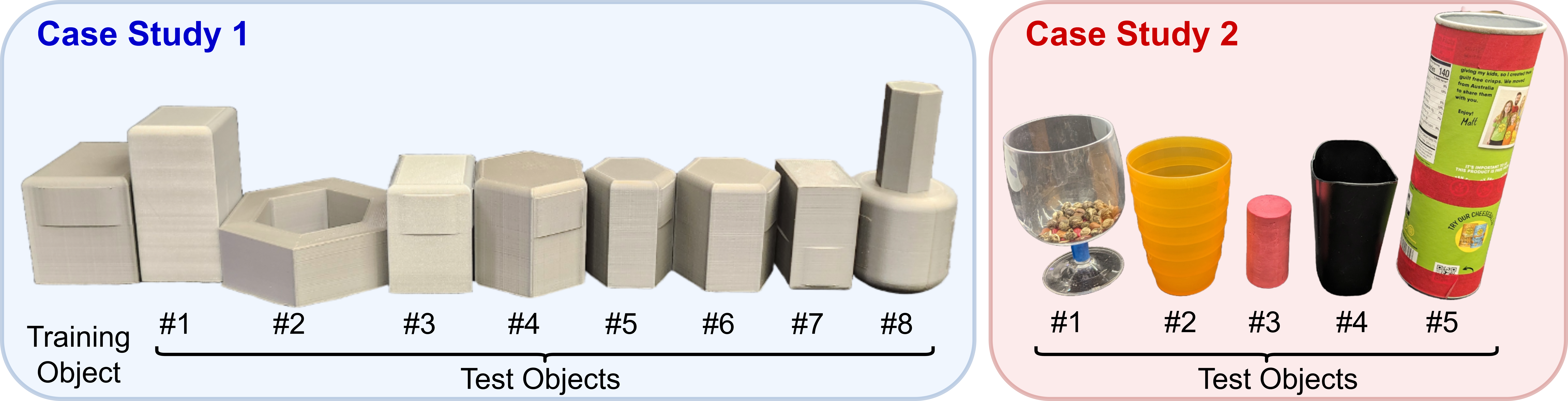}
    \caption{Solid 3D-printed objects with a smooth finish are used for tray-object transport experiments. In Case Study 1, a cube is exclusively used for collecting demonstrations, while other convex objects with varied inertial properties are used for testing. In Case Study 2, 5 poorest-performing objects from Case Study 1 are used for training, and five household objects with diverse materials and inertial properties are tested.}
    \label{fig:objects_tested}
\end{figure}

\noindent\textbf{Case Study 1 - Generalizing to Convex Shapes:}
We demonstrate the generalization capability of learned value functions to different convex shapes shown in Fig. \ref{fig:objects_tested}.
We train a value function ensemble using 50 demonstrations with a cube of mass 0.06 kg, a center of mass located 0.025 m above the contact surface, and a small friction coefficient ($\mu$) of 0.3, given the cube is 3D printed from PLA with a smooth finish. We evaluated the performance of CV-MPC on eight previously unseen convex objects with varying inertial and geometric properties (with same $\mu$). The value networks were not provided with the inertial and geometric properties of these test objects. We conduct three trials with different random seeds and randomly sampled goal locations. CV-MPC achieves a maximum success rate of 88.33\% (Tab. \ref{tab:gen1}). Interestingly, we observe that the learned value function utilizes tilting when necessary to maintain stability during object transport at sufficiently high velocities.\footnote{Refer to the website for experiment videos}

\begin{figure}%
    \centering    \includegraphics[width=0.85\linewidth]{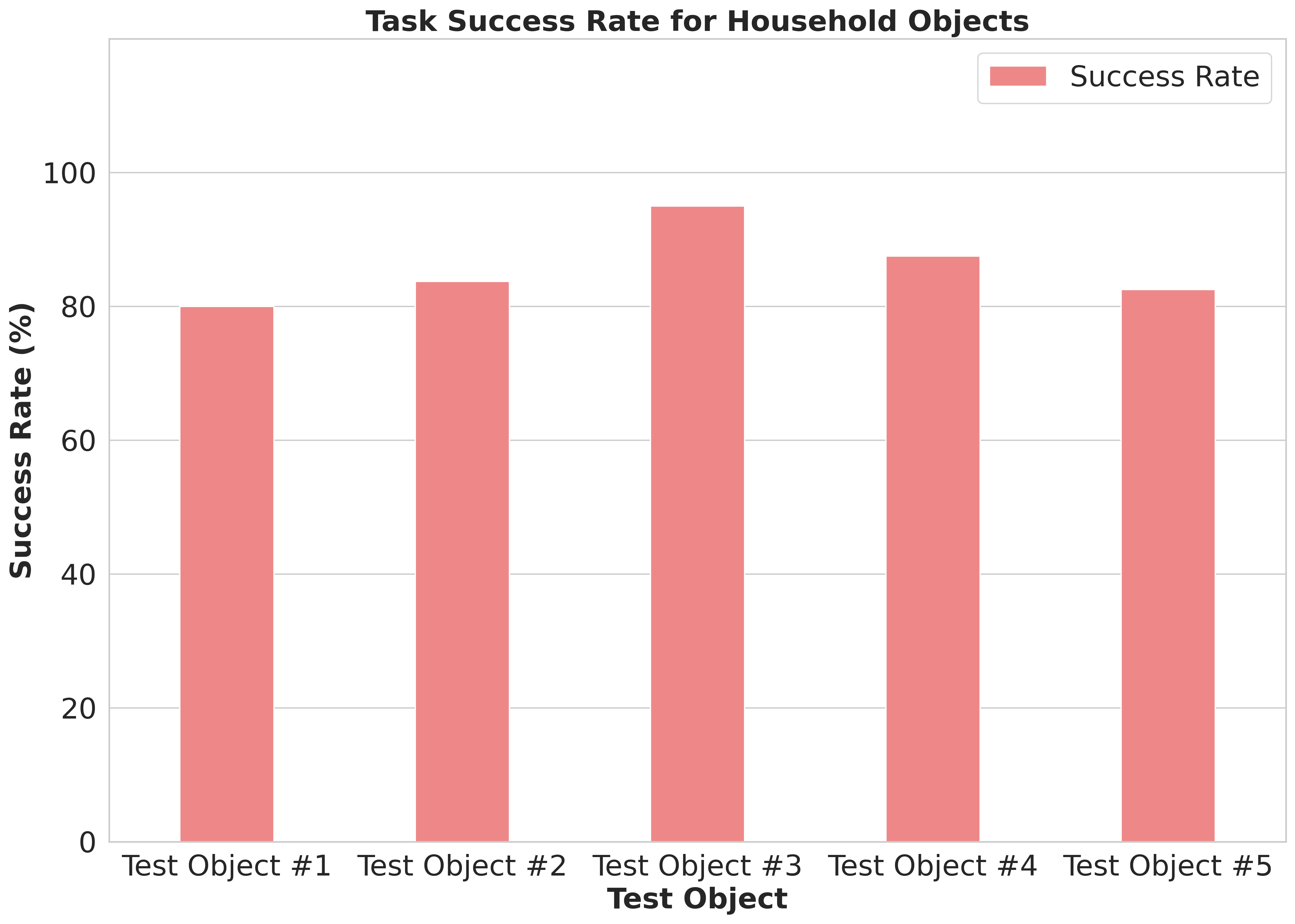}
    \caption{(Real-World Experiment) Success rates of CV-MPC on household objects of varied properties reported across 80 total trials per object. After training on 100 demonstrations from the five lowest-performing objects in Case Study 1, CV-MPC achieved high success rates on objects with entirely different shapes and materials, demonstrating its adaptability and efficacy.}
    \label{fig:household_obj}
\end{figure}

\begin{figure}%
    \centering
 \includegraphics[width=1.0\columnwidth]{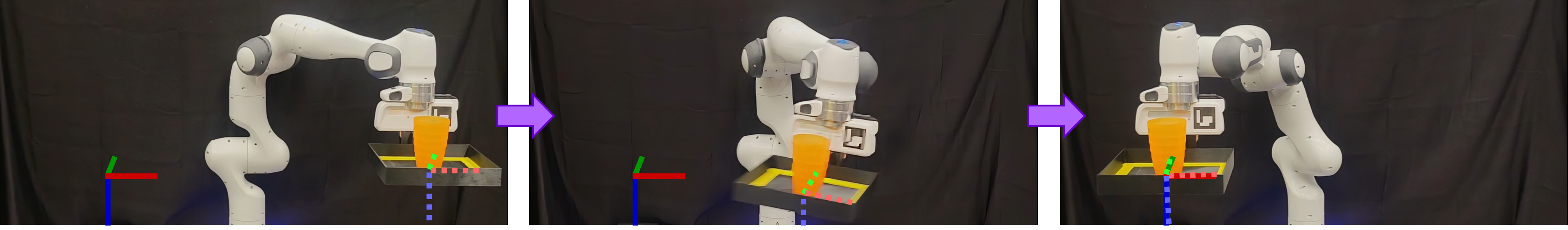}
    \caption{Franka Panda robot performing the harder lateral reaching task with an orange glass, which is one of the test objects in Case Study 2.}
    \label{fig:case_study3}
    \vspace{-0.1cm}
\end{figure}

\noindent\textbf{Case Study 2 - Generalizing to Household Objects:}
Next, we evaluate whether our approach generalizes to household objects with different geometries (Fig. \ref{fig:objects_tested}). We first tried the value function learned from the single cube and found that it did not lead to success on this object set. Next, we collected 20 demonstrations with 5 of the hardest objects 
from case study~1, totaling to 100 demonstrations. 
We conducted 80 trials per object, with 60 trails starting the robot near the center of the workspace as shown in Fig.~1. The remaining 20 trials had the robot start near the left of it's reachable workspace, with target positions near the right of it's reachable workspace as shown in Fig.~\ref{fig:case_study3}. 
As seen in Fig. \ref{fig:household_obj}, we observe that CV-MPC is able to generalize to various household objects and achieves a success rate of 80\% or above on all test objects, with the robot reaching, on average, a maximum linear velocity of 0.62 m/s, maximum angular velocity of 0.75 rad/s, and maximum tilt angle of 12.93 degrees throughout these trials.

\section{Discussion and Limitations}
\label{sec:discussion}
We presented a framework for teaching a robot manipulator the non-prehensile dynamic transport task from a small set of real-world demonstrations. We proposed a hybrid offline RL and MPC approach that enables rapid learning of new skills from a small amount of task-space demonstrations and easily integrates with existing off-the-shelf MPC frameworks to ensure robust deployment and generalization. We demonstrated the efficacy of our approach on learning the object transport task
with just 50-100 end-effector demonstrations in simulation and real-world while generalizing to novel objects at test time. However, there are a few key limitations. 
First, sampling-based MPC algorithms like the one used in this work~\cite{bhardwaj2022storm} are easy to use; however, ensuring hard constraints is challenging. An interesting direction for future work is to integrate different MPC optimizers that provide stronger guarantees on solution quality. Second, this work considers a perception free setting, however, integrating with pretrained perceptual models can further bolster generalization and performance. Finally, while CV-MPC shows great performance on the robot waiter task, it is essential to conduct a large-scale evaluation on a suite of dynamic manipulation tasks to test algorithmic performance. 

\section{Supplementary Material}
\subsection{Further Related Work}
\label{sec:related_work}
Our proposed approach for learning non-prehensile dynamic object transport draws upon several different research domains, and we provide a detailed overview of the related work in this section.\\

\noindent\textbf{Model-Predictive Control (MPC):} MPC is a powerful tool for feedback control of complex robotic systems under dynamic constraints with rich history of applications such as autonomous helicopter acrobatics~\cite{abbeel2010autonomous}, aggressive offroad driving~\cite{williams2016aggressive}, humanoid locomotion~\cite{erez_humanoidmpc_2013} and dynamic whole-body locomotion and manipulation~\cite{sleiman2021unified}. MPC approaches operate by solving for locally optimal policies for every state encountered during online execution by optimizing an approximate dynamics model and cost function over a finite planning horizon~\cite{wagener2019online}. We focus on a particular class of stochastic MPC algorithms that use sample-based estimates of policy gradient without restrictive assumptions on the class of policies, dynamics or cost functions. In recent times, many works have leveraged sampling-based MPC to enable real-world reactive manipulation~\cite{bhardwaj2022storm, pezzato2023samplingbased, jankowski2023vp}, with GPU acceleration and integration of learned components in the optimization loop. Our work builds on this framework and combines MPC with learned value functions from fixed demonstration data to enable sample-efficient learning of dynamic manipulation tasks.\\

\noindent\textbf{Learning for MPC:} The performance of MPC in practice can be limited by the quality of the dynamics model and the length of the optimization horizon. Several prior works propose data-driven methods to overcome these biases. For instance,~\cite{chua2018deep, williams2017information} learn a dynamics model from real-world interactions to correct for model bias. Further, ~\cite{erez2012infinite} and ~\cite{lowrey2018plan} propose learning value functions as terminal costs in MPC to increase the effective horizon and ~\cite{morgan2021model} use MPC as a policy class in an actor-critic setting. These approaches are often collectively referred to as model-predictive RL or infinite-horizon MPC.  Probably closest to ours is the method from~\cite{bhardwaj2020blending} that introduces a model-predictive Q-learning setup to simultaneously overcome model and finite-horizon biases by blending learned value estimates at all steps in the MPC horizon. However, most of these approaches are developed in the context of improving the sample efficiency of online RL and do not explicitly account for uncertainty due to missing data coverage when learning from fixed datasets. \\

\noindent\textbf{Batch Reinforcement Learning:}
Batch RL, also known as offline RL studies the problem of learning value functions and policies from fixed datasets of interactions generated by a behavior policy $\pi_{\beta}$ ~\cite{levine2020offline}. In many real-world scenarios, limited data coverage can lead to instabilities due to the covariate shift between the learner and behavior policy. Offline RL algorithms aim to systematically deal with limited data support by constraining the policy to stay away from states that have inadequate coverage. Leading approaches either employ a behavior regularization strategy~\cite{fujimoto2018addressing, wang2020critic}, where the policy is explicitly constrained to be close to the behavior policy, or a concept of pessimism-under uncertainty~\cite{kumar2020conservative, cheng2022adversarially} to optimize worst-case lower bounds on policy performance. In the latter setting, model-free algorithms operate by explicitly constraining the value function for out-of-distribution states and actions~\cite{kostrikov2021offline, kumar2020conservative}, and optimizing the policy with respect to such a conservative value function. Our proposed approach for offline value function learning and online conservative MPC bears close resemblance to this setting and is especially related to recent algorithms that learn an ensemble of value functions to construct performance lower bounds~\cite{an2021uncertainty, nikulin2022q}. However, while these approaches learn a policy offline, we integrate the learned value functions with online MPC using a conservative estimate of trajectory returns. Another recent approach to note is from ~\cite{xie2021bellman} that shows how many offline RL algorithms based on value function pessimism can be overly pessimistic by constraining the learned value function at every state (pointwise pessimism) and proposes a scheme to overcome this by inducing pessimism in the value function only at the start state (initial-state pessimism). This is related to our conservative online MPC, where we construct pessimistic (or conservative) returns from the value function only at the first state and do not constrain value predictions to be pessimistic at every intermediate state in MPC rollouts. 

Finally, there have been several recent works like~\cite{sikchi2022learning, argenson2020model} that combine offline RL with online MPC and have shown strong performance in simulated benchmarks. However, key differences in our approach are (1) learning from only end-effector demonstrations, (2) using initial state pessimism instead of point-wise pessimism as in the above approaches, and (3) a practical, real-world deployment by leveraging powerful off-the-shelf MPC optimizers.\\

\noindent\textbf{Learning from Observations Alone:} Learning policies from observation-only demonstrations without knowledge of expert actions is an active area of research, with both model-based and model-free approaches. In the model-based setting, ~\cite{torabi2018behavioral} propose to learn an inverse dynamics model from a small dataset of unsupervised environment interactions and use it to infer expert actions from given demonstrations. This is followed by learning a policy via behavior cloning on the inferred actions.~\cite{pavse2020ridm} introduce a similar inverse dynamics learning approach. However, the model is trained to maximize task reward rather than trajectory tracking error. While our proposed CV-MPC is also model-based, unlike these methods, it does not require further environment interactions to learn inverse dynamics models but rather relies on powerful MPC algorithms to optimize joint space motions. In the model-free domain,~\cite{sun2019provably} provides a provably efficient algorithm for imitation from observations by matching the expert's next state distribution. Similarly, ~\cite{torabi2018generative} take a distribution matching perspective and formulate an adversarial training approach for learning from observations alone akin to Generative Adversarial Imitation Learning (GaIL)~\cite{ho2016generative}. Related is the work from~\cite{kimura2018internal} that learns a model to predict the expert's next state that is used as a reward for policy optimization. However, similar to the model-based approaches before, these require online interactions with the system during learning.~\cite{li2023mahalo} generalize this to the setting of offline RL and provide strong theoretical guarantees but rely on a separately collected dynamics dataset for optimizing policies.  

\subsection{Algorithmic Demonstrator using MPC with Friction Cone Constraints}
\label{sec:mpcdemonstrator}
While our approach does not make any assumptions about the source of demonstrations, in this work, we use an algorithmic demonstrator, which enables us to collect demonstrations for different ablation studies easily. We use an MPC-based demonstrator that uses friction cone constraints to enforce object stability, building on the formulation described in~\cite{heins2023keep}.  

The object, denoted as $O$, is modeled as a rigid body adhering to the Newton-Euler equation ${w}_{GI}$ + ${w}_C$ $=$ 0 where ${w}_{GI}$ is the gravitoinertial wrench and ${w}_C$ is the contact wrench.
Our aim is to create an expert that implicitly satisfies the Newton-Euler equation as well as the friction cone constraints so that the object doesn't slide substantially while reaching the target goal. For this, we only assume access to robot states and nominal values for object friction and inertial properties. 
We use these dynamics equations to formulate a cost function that only penalizes end-effector states that violate the constraint and is zero elsewhere. The gravitoinertial wrench is given by

\begin{equation}
\label{eq:wgi}
\begin{aligned}
w_{GI} = \begin{bmatrix} f_{GI} \\ \tau_{GI} \end{bmatrix} &= - \begin{bmatrix} m (\dot{v}_o - R_o g) \\ J \dot{\omega}_o + \omega_o \times J \omega_o \end{bmatrix} \\
&= - \begin{bmatrix} m (\dot{v}_e - R_e g) + m (\omega_e^{\times} + \omega_e^{\times} \omega_e^{\times}) c \\ J \dot{\omega}_e + \omega_e^{\times} J \omega_e \end{bmatrix}
\end{aligned}
\end{equation}
\noindent
where ${f}_{GI}$ and ${\tau}_{GI}$ are the gravitoinertial force and torque, respectively, $m$ is the object mass, ${v}_o$ and ${\omega}_o$ are the body-frame linear and angular velocities of the object center of mass (CoM), ${g}$ is the gravitational acceleration, and ${J}$ is the object inertia matrix about the CoM. The rotation matrix ${R}_o$ represents the object's orientation with respect to the world and is used to rotate gravity into the body frame. Similarly, ${v}_e$ and ${\omega}_e$ are the end-effector frame linear and angular velocities, ${R}_e$ is the rotation matrix for the end-effector frame, and ${c}$ is the position of the object's CoM with respect to the end effector. The notation $(.)^{\times}$ denotes the skew-symmetric conversion operator. Here, the first equality is the definition of gravitoinertial wrench with velocities and orientation expressed in the object's body frame. The second equality results from the assumption that the object moves minimally on the tray, thus allowing us to re-write the Newton-Euler equations to the end-effector frame by approximating ${R}_{o} \approx {R}_e$. 

As mentioned before, to ensure the object does not slip on the tray surface, we require the gravitoinertial wrench $w_{GI}$ to be balanced by the contact wrench $w_{C}$ in the object body frame $\lbrace B \rbrace$. To model the body wrench resulting from the object being in contact with the tray, we
calculate the forces at pre-defined contact points on the object surface using point contact with friction model as used in ~\cite{selvaggio2022nonpreobjtransport}. Given $n$ contact points, let $F_C$ denote the stacked contact force vector ${F}_C = 
\begin{bmatrix}
{f}_{c_1}^T & \ldots & {f}_{c_{n}}^T
\end{bmatrix}^T \in \mathbb{R}^{3n}$ where each ${f}_{c_i}$ consists of tangential and normal force components at the $i$-th contact point. 
$F_C$ can be related to the gravitoinertial wrench using the Grasp matrix $G$ as 
\begin{equation}
\label{eq:grap_mat}
{F}_C = {G}^{-1} {w}_{GI}
\end{equation}
\noindent
where ${G}$ is defined as ${G} = 
\begin{bmatrix}
\text{Ad}_{{T}_{{{q}_{b,c}}^{-1}}} {B}_{c,1} & \ldots & \text{Ad}_{{T}_{{{q}_{b,c}}^{-1}}}{B}_{c,n}
\end{bmatrix}$. Here, ${G}^{-1}$ denotes the pseudo-inverse ${G}$, $\text{Ad}_{{T}_{{{q}_{b,c}}^{-1}}}$ represents the adjoint transformation matrix that relates the contact point ${c}_i$ to object frame \{$B$\} such that ${q}_{b,c}$ is the contact point pose in \{$B$\}, and ${B}_{c,i}$ represents the basis matrix that projects the components of the contact forces that are transmissible through ${c}_i$ into 6D space.  The grasp matrix ${G}$ thus encapsulates the relationship between the individual contact forces and the resultant wrench acting on the object.
Given this relationship, we can compute the contact forces $F_C$ resulting from end-effector motions. 
To ensure no slip, the contact forces at each contact point ${f}_{c_i}$ must satisfy the friction cone constraints:
\begin{equation}
Constraint = \left\{ {f}_{c_i} \in \mathbb{R}^3 
\begin{array}{l}
    : \sqrt{f_{c_i,x}^2 + f_{c_i,y}^2} \leq \mu f_{c_i,z} \\
    , \; f_{c_i,z} \geq 0 
\end{array}
\right\}
\end{equation}
To optimize robot trajectories that satisfy this constraint, we formulate a cost $\hat{G}_{friction}$ as
\begin{equation}
\hat{c}_{friction} =
\begin{cases}
\sqrt{f_{c_i,x}^2 + f_{c_i,y}^2} - \mu f_{c_i,z} & 
\begin{array}{@{}l@{}}
    \text{if } \sqrt{f_{c_i,x}^2 + f_{c_i,y}^2}\\
    > \mu f_{c_i,z}, \; f_{c_i,z} \geq 0
\end{array}\\

0 & \text{otherwise}
\end{cases}
\end{equation}
In our implementation, we integrate this as a part of the running cost in the \storm framework, allowing us to collect demonstrations for different experiments. However, note that while the algorithmic demonstrator assumes access to object inertial and friction properties, the learned value function does not. In our experiments, we also study how learning can improve upon suboptimal demonstrations when the nominal object properties are incorrect. 

\begin{figure*}
    \centering
    \begin{subfigure}[b]{0.25\textwidth}
        \centering
        \includegraphics[width=0.8\textwidth]{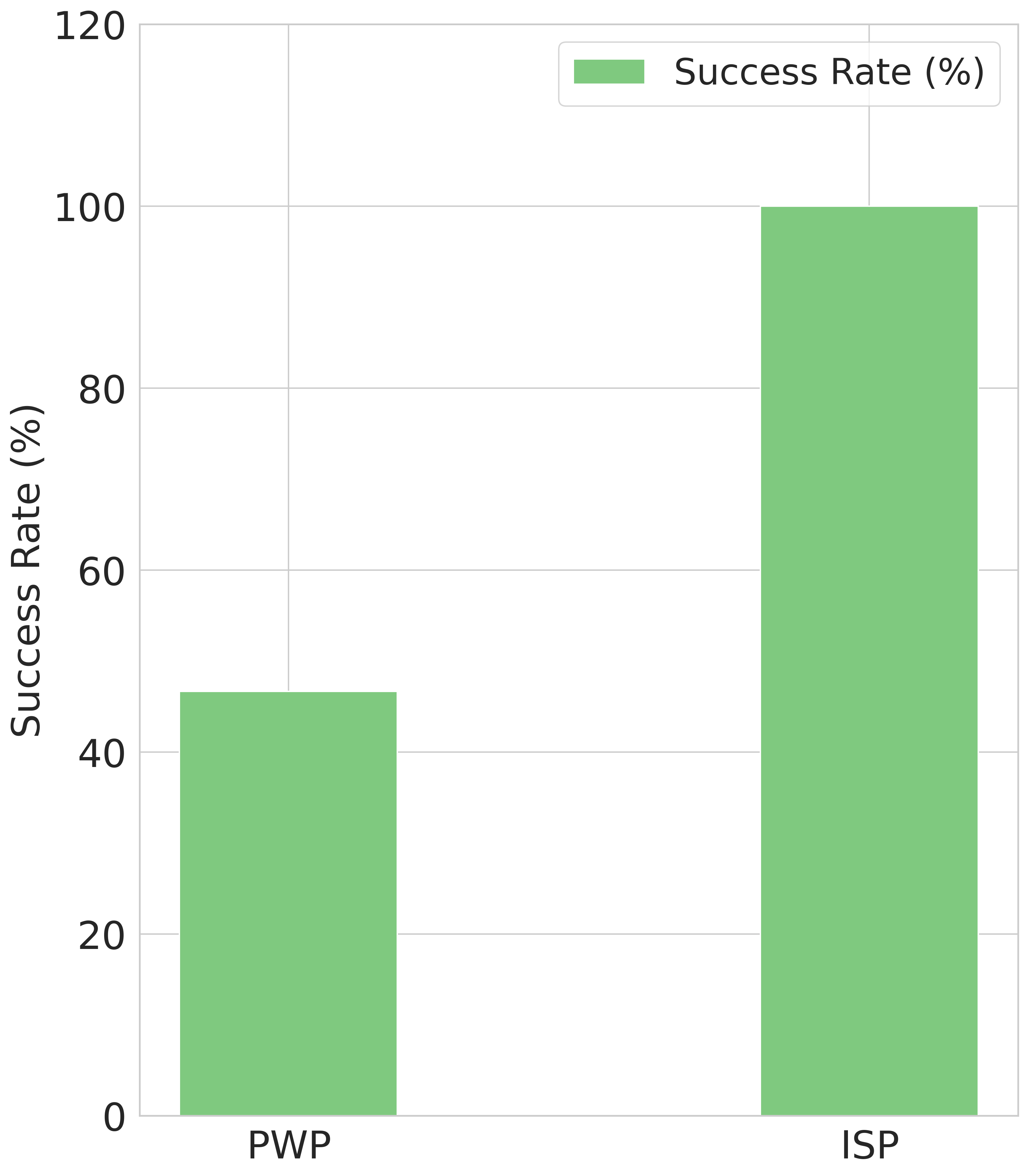}
        \label{fig:success_rates}
    \end{subfigure}
    \hspace{-0.2cm} %
    \begin{subfigure}[b]{0.25\textwidth}
        \centering
        \includegraphics[width=0.8\textwidth]{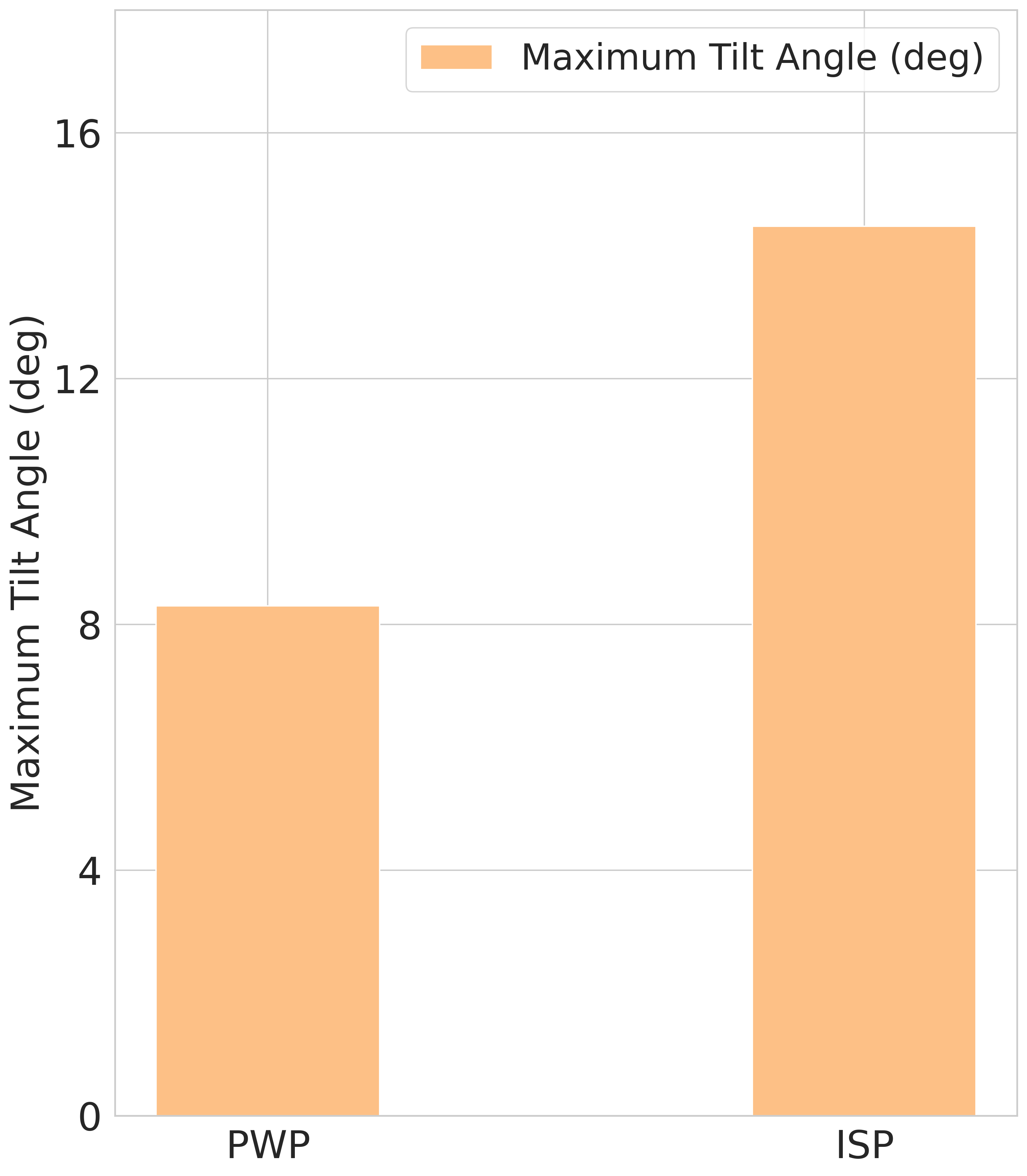}
        \label{fig:alpha}
    \end{subfigure}
    \hspace{-0.2cm} %
    \begin{subfigure}[b]{0.25\textwidth}
        \centering
        \includegraphics[width=0.8\textwidth]{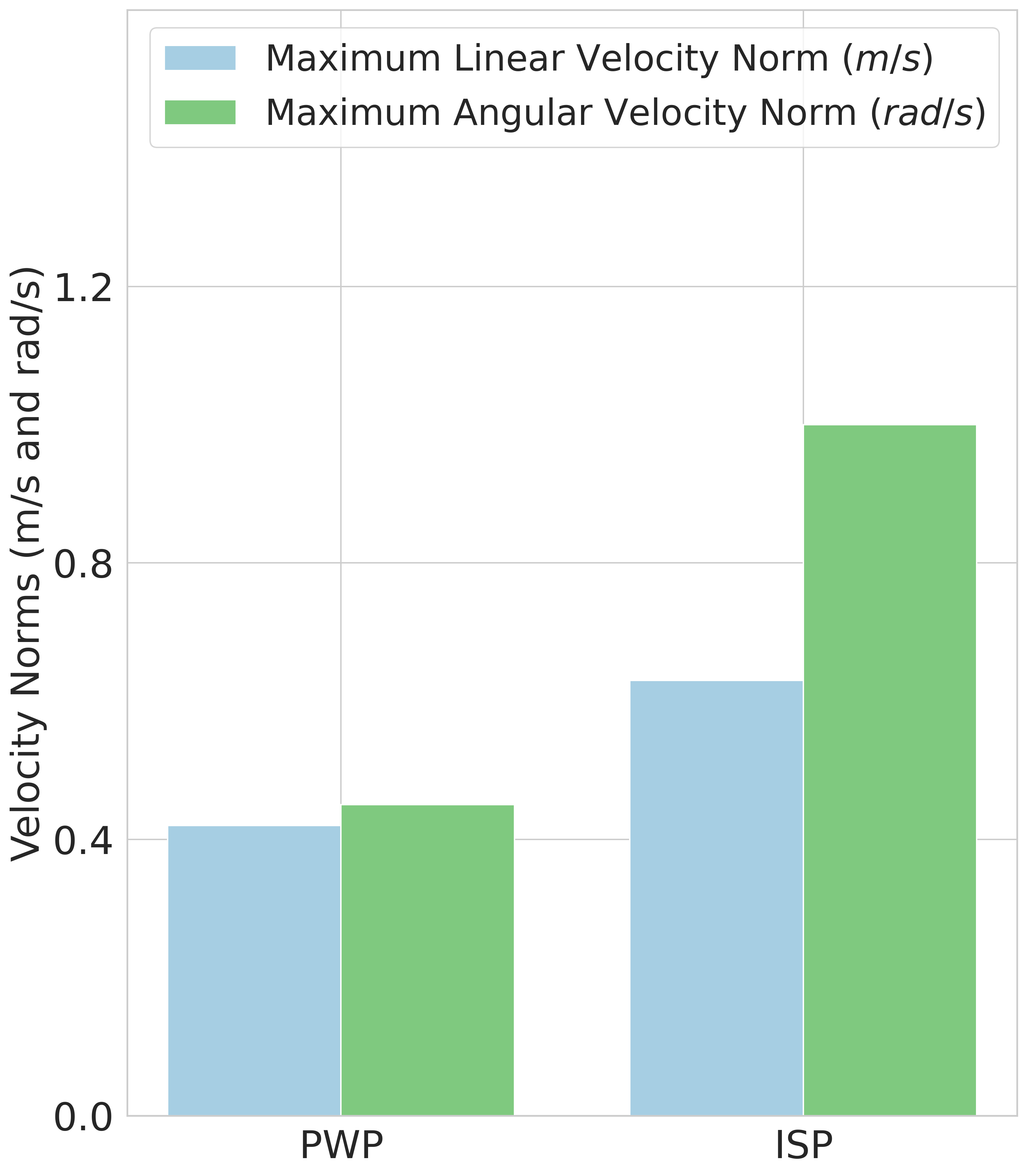}
        \label{fig:vels}
    \end{subfigure}
    \vspace{-0.1cm}
    \caption{(Simulation Experiment) Performance ablation comparing Pointwise Pessimism (PWP in the figure) and Initial-State Pessimism (ISP in the figure) as value function blending schemes. Over-pessimism in the former results in reduced success rates and less dynamic motion compared to the latter.}
    \label{fig:pwp_isp}
    \vspace{-0.2cm}
\end{figure*}

\subsection{Further Ablations}
\subsubsection{Comparing Observations Spaces}

Since our approach involves only learning value functions for stabilizing behavior during non-prehensile object transport, we found that we do not need to condition the learned value function on the end-effector targets. We hypothesize that the end-effector rotation ($R_e$) is the most critical observation for the value functions to learn effectively and generalize across different start locations. To validate this hypothesis, we conducted ablation studies under two conditions: using the same end-effector start location for both training and testing and using different start locations for training and testing, and examine four sets of observations: \\

\textbf{Full Observation} ($o^{\text{full}}_t$): Including position, velocity, acceleration, and rotation of the end-effector.
\[
    o^{\text{full}}_t = x^{\text{full}}_t = \begin{bmatrix}
    p^{ee}_t & v^{ee}_t & a^{ee}_t & R_e
    \end{bmatrix}^T
\]

\textbf{Velocity, Acceleration, and Rotation} ($o^{\text{vel\_acc\_rot}}_t$): Omitting position but including velocity, acceleration, and rotation.
\[
    o^{\text{vel\_acc\_rot}}_t = x^{\text{vel\_acc\_rot}}_t = \begin{bmatrix}
    v^{ee}_t & a^{ee}_t & R_e
    \end{bmatrix}^T
\]

\textbf{Velocity and Acceleration} ($o^{\text{vel\_acc}}_t$): Including only velocity and acceleration.
\[
    o^{\text{vel\_acc}}_t = x^{\text{vel\_acc}}_t = \begin{bmatrix}
    v^{ee}_t & a^{ee}_t
    \end{bmatrix}^T
\]

\textbf{Rotation Only} ($o^{\text{rot}}_t$): Including only rotation.
\[
    o^{\text{rot}}_t = x^{\text{rot}}_t = R_e
\]

By systematically varying these observation sets, we aim to understand how different end-effector observations affect the performance of the value functions, thereby testing our hypothesis.\\

\begin{figure}[H]
    \centering
    \includegraphics[width=0.55\linewidth]{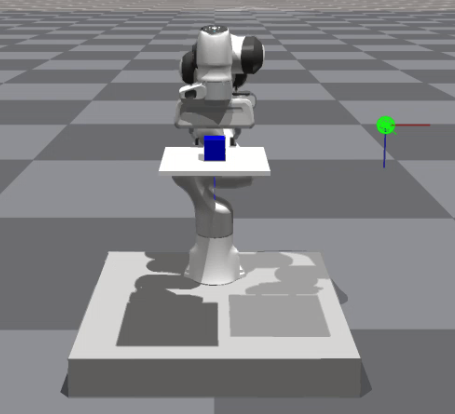}
    \caption{Snapshot of simulation ablation for different observation spaces with the same EE start position.}
    \label{fig:cube_center_sim}
\end{figure}

\noindent\textbf{a) Same Start EE Position at Train and Test - Generalizing to Different Goal Positions (Easy)}: In this experiment, we collected 50 demonstrations in a simulation with the cube positioned at the center as shown in Fig \ref{fig:cube_center_sim}. The same start location was used for both all training episodes as well as testing, while goals were randomly sampled. The experiment was tested across 60 episodes. The results given in Table ~\ref{tab:same_loc} indicate that the observation sets $o^{\text{full}}_t$, $o^{\text{vel\_acc\_rot}}_t$, and $o^{\text{rot}}_t$ all achieved a success rate of 100\%, whereas the observation set $o^{\text{vel\_acc}}_t$ yielded a success rate of only 5\%. It is noteworthy that $o^{\text{vel\_acc}}_t$ is the only set that does not include $R_e$ as an observation for the value functions, thereby confirming our hypothesis.\\

\vspace{-0.5cm} %

\begin{table}[H]
    \centering
    \renewcommand{\arraystretch}{2}
    \begin{adjustbox}{max width=\linewidth}
    \begin{tabular}{|c|
    >{\centering\arraybackslash}m{2.5cm}|>{\centering\arraybackslash}m{2.5cm}|>{\centering\arraybackslash}m{2.5cm}|>{\centering\arraybackslash}m{2.5cm}|}
    \hline
    \textbf{Metrics:} & 
    \textbf{$o^{\text{full}}_t$} & 
    \textbf{$o^{\text{vel\_acc\_rot}}_t$} & 
    \textbf{$o^{\text{vel\_acc}}_t$} & 
    \textbf{$o^{\text{rot}}_t$} \\
    \hline \textbf{$Success$ (\%)} &  100.00 & 100.00 & 5.00 & 100.00 \\ 
    \hline \textbf{${\alpha_{\mathrm{ee}, t}^{\max} }$ (deg)} & 14.58 ± 0.38 & 14.64 ± 0.46 & 10.99 ± 1.93 & 13.39 ± 0.39 \\ 
    \hline \textbf{$\|\mathbf{v}_{\mathrm{ee}, t}\|_{\max}$ (m/s)} & 0.63 ± 0.04 & 0.58 ± 0.03 & 0.47 ± 0.05 & 0.62 ± 0.03 \\ 
    \hline \textbf{$\|\boldsymbol{\omega}_{\mathrm{ee}, t}\|_{\max}$ (rad/s)} & 1.00 ± 0.05 & 0.99 ± 0.05 & 0.56 ± 0.04 & 0.89 ± 0.01\\ 
    \hline
    \end{tabular}
    \end{adjustbox}
    \caption{Ablation study on end effector observations to the value function networks, highlighting that $R_e$ plays a critical role in successful value function learning.}
    \label{tab:same_loc}
\end{table}

\begin{figure}[H]
    \centering
    \includegraphics[width=0.55\linewidth]{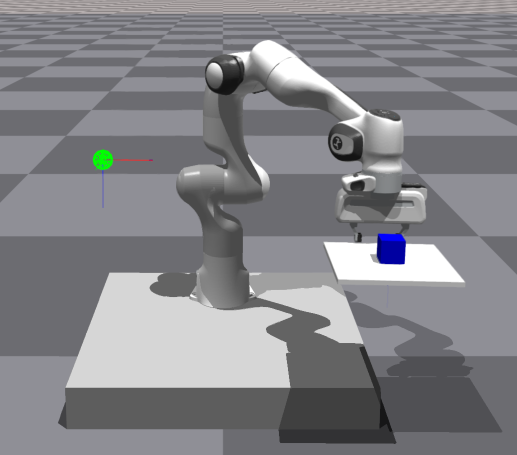}
    \caption{Snapshot of simulation ablation for different observation spaces with varying EE start positions during train and test.}
    \label{fig:sim_diff_cube_pos}
\end{figure}

\noindent\textbf{b) Different Train and Test Start EE Positions - Generalizing to Different Goal Locations (Difficult)}: To further assess the generalizability of the value functions and their sensitivity to different observations, we conducted an ablation study to determine which observation set leads to better generalization with respect to different starting positions. For this purpose, we used the same value functions trained from the demonstrations in the previous ablation. During inference, we employed different initial positions of the robot than those used during training, as seen in Fig \ref{fig:sim_diff_cube_pos}. As shown in Table~\ref{tab:diff_start_loc}, the observation set $o^{\text{rot}}_t$ exhibited the best performance with a success rate of 66.67\%. The observation set $o^{\text{vel\_acc\_rot}}_t$ also demonstrated satisfactory performance with a success rate of 56.67\%. As expected, the observation set $o^{\text{vel\_acc}}_t$ performed the worst, as rotation is a critical aspect of stability. Additionally, the observation set $o^{\text{full}}_t$ showed poor performance since the end-effector position during training differed from that during testing, complicating generalization based on position. 

\vspace{-0.9cm} %

\begin{table}[H]
    \centering
    \renewcommand{\arraystretch}{2}
    \begin{adjustbox}{width=\linewidth}
    \begin{tabular}{|c|
    >{\centering\arraybackslash}m{2.5cm}|>{\centering\arraybackslash}m{2.5cm}|>{\centering\arraybackslash}m{2.5cm}|>{\centering\arraybackslash}m{2.5cm}|}
    \hline
    \textbf{Metrics:} & 
    \textbf{$o^{\text{full}}_t$} & 
    \textbf{$o^{\text{vel\_acc\_rot}}_t$} & 
    \textbf{$o^{\text{vel\_acc}}_t$} & 
    \textbf{$o^{\text{rot}}_t$} \\
    \hline \textbf{$Success$ (\%)} &  28.33 & 56.67 & 5.00 & 66.67 \\ 
    \hline \textbf{${\alpha_{\mathrm{ee}, t}^{\max} }$ (deg)} & 14.40 ± 0.41 & 16.13 ± 0.21 & 10.97 ± 0.37 & 12.55 ± 0.89 \\ 
    \hline \textbf{$\|\mathbf{v}_{\mathrm{ee}, t}\|_{\max}$ (m/s)} & 0.46 ± 0.02 & 0.52 ± 0.01 & 0.44 ± 0.03 & 0.58 ± 0.01 \\ 
    \hline \textbf{$\|\boldsymbol{\omega}_{\mathrm{ee}, t}\|_{\max}$ (rad/s)} & 0.66 ± 0.02 & 0.58 ± 0.04 & 0.56 ± 0.04 & 0.81 ± 0.01\\ 
    \hline
    \end{tabular}
    \end{adjustbox}
    \caption{Ablation study on end effector observations to the value function networks, showing that $R_e$ alone outperforms other end effector observations in terms of success, as anticipated from the previous experiment.}
    \label{tab:diff_start_loc}
\end{table}

\subsubsection{Pointwise Pessimism vs. Initial-State Pessimism}
We aim to investigate the effect of the initial-state pessimism scheme in comparison to the pointwise pessimism scheme for handling sparse reward data from demonstrations~\cite{xie2021bellman}. For this, we train an ensemble of value functions from 50 demonstrations using the same cube at the center and a similar experimental setup used in previous ablations. During inference, we compute a conservative value estimate (as per Eq.~\ref{eq:pess_returns} in the main text) instead of inducing pessimism estimates at all intermediate rollout steps. We hypothesize that excessive pessimism can lead to motions that may hinder performance in challenging dynamic tasks such as the robot waiter problem. This hypothesis is supported by the results shown in the bar plots in Figure~\ref{fig:pwp_isp}. Due to over-pessimism, the former scheme results in a success rate of less than 50\%, whereas the latter achieves a success rate of 100\%. Additionally, we observe less dynamic behavior due to over-pessimism in the former approach. Specifically, the maximum tray tilt angle and velocities are lower in successful episodes for the former compared to the latter blending scheme.

\bibliographystyle{IEEEtran}
\bibliography{ref}

\end{document}